\pdfoutput=1

\documentclass{article}

\usepackage[preprint]{neurips_2026}

\usepackage[table]{xcolor}

\definecolor{tblHeader}{RGB}{245,247,250}
\definecolor{tblStripe}{RGB}{250,250,252}

\definecolor{catMLE}{RGB}{235,242,255}   
\definecolor{catNLP}{RGB}{235,255,240}   
\definecolor{catCV}{RGB}{255,240,235}    
\definecolor{catTab}{RGB}{245,235,255}   

\definecolor{lightHR}{gray}{0.95}
\definecolor{liteBlue}{RGB}{240,245,255}
\definecolor{liteGreen}{RGB}{240,248,240}
\definecolor{liteOrange}{RGB}{255,248,235}
\definecolor{litePurple}{RGB}{245,240,250}

\usepackage[utf8]{inputenc} 
\usepackage[T1]{fontenc}    
\usepackage{hyperref}       
\usepackage{url}            
\usepackage{booktabs}       
\usepackage{amsfonts}       
\usepackage{nicefrac}       
\usepackage{microtype}      
\usepackage{xcolor}       
\usepackage{xspace}
\usepackage{wrapfig}
\usepackage{graphicx}
\usepackage{enumitem}
\usepackage[table]{xcolor}
\usepackage{array}         
\usepackage{booktabs}      
\usepackage{longtable}
\usepackage{booktabs}
\usepackage{array}
\usepackage{ragged2e}
\usepackage{amsmath, amssymb}

\newcommand{\eg}{\emph{e.g.}}
\newcommand{\method}{\texttt{Matryoshka Agent}\xspace}
\definecolor{nblue}{cmyk}{0.95,0.0,0.2,0.2}

\definecolor{morandiBlue}{RGB}{236,240,244}
\definecolor{morandiGreen}{RGB}{235,240,237}
\definecolor{morandiGray}{RGB}{245,246,247}

\title{\method: Unfolding Sub-Agents for Long-Horizon Machine Learning Engineering}

\author{%
  \textbf{Rushi Qiang} \quad
  \textbf{Changhao Li}\thanks{Equal second authorship.} \quad
  \textbf{Haotian Sun}\footnotemark[1] \\
  \textbf{Yuchen Zhuang} \quad \textbf{Chao Zhang} \quad \textbf{Bo Dai} \\
  Georgia Institute of Technology \\
  \texttt{\{rqiang6, cli911, haotian.sun\}@gatech.edu} \\
  \texttt{night.yuchen@gmail.com} \quad
  \texttt{chaozhang@gatech.edu} \quad
  \texttt{bodai@cc.gatech.edu} \\
}

\begin{document}

\maketitle

\begin{abstract}



Machine learning engineering (MLE) tasks require long-horizon decision making over iterative solution debugging and refinement, under expensive and feedback-driven environment interactions.
Developing and training a monolithic agent for such tasks is fundamentally challenging, as it must simultaneously manage extremely long and noisy contexts, explore vast solution spaces, and remain effective under limited model capacity and computational budgets.
To address these challenges, we propose \method, a unified hierarchical agent framework for complex long-horizon tasks.
\method decomposes agentic problem solving into a coordinated hierarchy of decision making and execution: a high-level Orchestrator maintains compact, long-horizon exploration states and issues strategic instructions, while lower-level Sub-Agents execute concrete solution attempts through direct environment interaction, mediated by standardized Tool interface.
This design decouples strategic exploration from costly execution, substantially reducing the burden of long-context reasoning and enabling efficient iterative refinement.
We further develop an efficient training paradigm
for \method.
Experimental results on a broad range of MLE tasks with diverse model types and scales demonstrate that \method is an effective and scalable paradigm for long-horizon MLE tasks and complex agentic problem solving. Notably, \method enables Qwen3-4B-Instruct to reach Orchestrator performance comparable to o4-mini. 
Applying \method to Qwen3-30B-Coder results in at most 36.7\% relative performance gain.

\end{abstract}

\section{Introduction}



Large Language Models (LLMs) have developed from static question-answering (QA) chatbots to autonomous agents that are capable of solving complex, multi-step challenges through interaction with real-world environments~\cite{chan2024mle,jimenez2023swe,qiang2025mledojo,zhou2023webarena,merrill2026terminal,wei2025browsecomp,yao2024tau,xie2024osworld,patwardhan2025gdpval}. 
Within this expanding landscape, a new class of performance-optimization agentic tasks has emerged, where the goal is not merely to complete a single executable objective, but to progressively improve solution quality through feedback from the environment~\cite{ouyang2025kernelbench,press2025algotune,rank2026posttrainbench}. These tasks are typically characterized by several fundamental challenges. 
First, long-horizon interaction naturally produces growing contexts, with execution logs, intermediate outputs, and debugging feedback accumulated across rounds~\cite{du2025deepresearch,wei2025browsecomp}.
Second, these tasks involve large, open-ended action and observation spaces, where diverse observations can induce a broad range of subsequent refinement choices~\cite{phan2025humanity,merrill2026terminal,ouyang2025kernelbench,press2025algotune}.
Third, environment interactions often incur substantial wall-clock and computational cost, making both inference-time exploration and training-time data collection expensive~\cite{starace2025paperbench,huang2023mlagentbench,rank2026posttrainbench}.
Machine Learning Engineering (MLE) tasks, characterized by their dependency on task-specific datasets with diverse modalities (\eg, images, text), schemas, data distributions, and evaluation objectives, constitute a particularly challenging and representative instantiation of these difficulties~\cite{nam2025mle,jing2024dsbench,chan2024mle,qiang2025mlesmith,huang2023mlagentbench}.
Consequently, effective MLE agents must combine strong coding and debugging skills with deep domain understanding and the ability to systematically explore and exploit diverse methodological alternatives~\cite{jiang2025aide,li2024autokaggle,yang2025r,guo2024ds,nam2025mle,toledo2025ai}, which inevitably requires accumulating extensive history contexts that include redundant information. 
Furthermore, the practical execution of MLE code is often computationally intensive and time-consuming, especially for large-scale datasets. The need to handle long contexts under noisy and partially redundant histories, the substantial time and computational costs of environment interaction, and the requirement for sustained exploration and iterative refinement jointly pose significant challenges for agents to effectively solve this important yet highly complex class of tasks.

In response to these challenges, we propose \method, a hierarchical framework for long-horizon
MLE tasks. \method separates iterative performance optimization into three coordinated layers. Orchestrator serves as the high-level decision layer: it maintains a compact cross-round state over previous instructions, summarized outcomes, and evaluation scores, and decides which
solution direction to refine next.
Sub-Agents serve as the execution layer: they are instantiated with fresh contexts to implement, run, and debug concrete solution attempts through direct environment interaction.
Tools mediate between these layers by formatting Orchestrator instructions, injecting selected references from previous attempts, launching Sub-Agents, and returning structured summaries and
scores. This design creates an efficient separation between strategic evolution and concrete execution, reducing long-context burden while preserving key information for iterative refinement.
Beyond the inference-time framework, \method further develop a training paradigm tailored to expensive performance-optimization tasks.
During training, Solution Refinement Tree--based sampling reuses shared execution prefixes and collects structured comparisons among alternative refinements from common states.
These comparisons provide trajectory-level preference signals for optimizing the Orchestrator with online ranking-based reinforcement learning, while successful execution trajectories being
used to improve Sub-Agents.

\begin{figure}
    \centering
    \includegraphics[width=\linewidth]{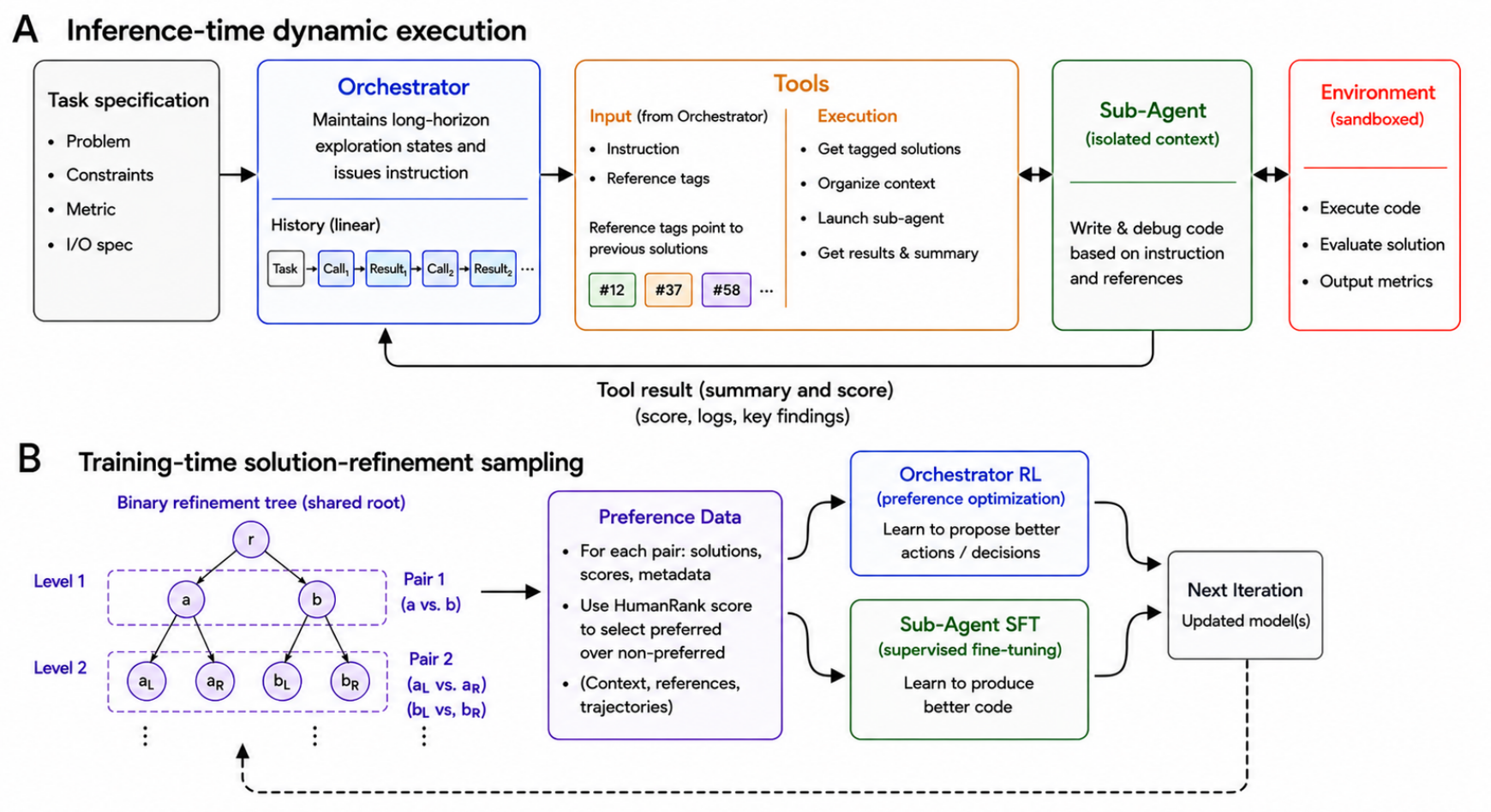}
    \caption{Overview of \method. Upper: Interaction Logic among the three-layered structure. Lower: Training Paradigm with Solution Refinement Tree--based sampling.}
    \label{fig:main-fig}
    \vspace{-3ex}
\end{figure}
Our contributions can be summarized as follows:
\begin{itemize}[leftmargin=1.2em, itemsep=0pt, topsep=2pt]
    \item We propose \method, a hierarchical MLE agent framework that decouples high-level
    orchestration from low-level code execution through structured tool-
    mediated feedback.
    \item We introduce a training paradigm based on Solution Refinement Tree--based sampling and
    trajectory-level ranking optimization, enabling efficient Orchestrator learning from costly
    environment interactions while supporting Sub-Agent improvement from successful executions.
    \item We evaluate and analyze \method extensively across a wide range of MLE tasks with diverse model types, scales, and combinations. Experimental results demonstrate the effectiveness of both \method and the training paradigm.
\end{itemize}

\section{Related Works}

\paragraph{Modularized LLM Agents}
Recent advances in LLM-based agents increasingly emphasize modular architectures, where specialized components coordinate to handle diverse decision-making contexts~\cite{wang2024simulating, hong2024metagpt, yan2025selftalk, chen2024autoagents, zhang2024chainagents}. Interactions among such modules have shown strong potential for improving reasoning, exploration, and task execution~\cite{liang2024encouraging, talebirad2023multi, chen2023agentverse, bettini2024benchmarl, zhang2026agentorchestra}.
Early modular frameworks primarily rely on advisory or controller-based guidance by introducing trainable modules that steer frozen foundation models through intermediate signals~\cite{asawa2025train, shang2025agentsquare, wei2025reactplannercentric}. Moving beyond this paradigm, Matryoshka-Pilot~\cite{li2025matryoshka} adopts a controller–generator framework, where a lightweight trainable controller explicitly controls a black-box LLM via structured guidance.
Subsequent work explores explicit decomposition into specialized planning, execution, verification, and generation modules, while maintaining a subset of trainable components to improve efficiency~\cite{chen2025agentflow, zhang2026agentorchestra, wan2025rema}.
More recent approaches adopt a subagent-as-tool paradigm, enabling submodules to co-evolve through self-generated tasks and rewards via dual-role and multi-role self-play interactions~\cite{jiang2025adaptation, huang2025r, chen2025multi}.
Despite their promise, modular agent systems face substantial challenges in system complexity, training cost, and stability~\cite{cemri2025multi, zhang2025lazyagent, yan2025selftalk}. Extending the research to long-horizon agentic workflows, we propose \method, a modular LLM agent framework that systematically orchestrates end-to-end solution pipelines and achieves strong performance on complex MLE tasks.

\paragraph{Coding Agents for MLE}
Leveraging rapid advances in LLM coding and reasoning capabilities~\cite{jain2024livecodebench, jimenez2023swe}, recent work has explored LLM-based coding agents for MLE tasks. These efforts span agentic solutions for ML tasks~\cite{chan2024mle, qiang2025mledojo, qiang2025mlesmith, tang2023ml, huang2023mlagentbench}, data science problems~\cite{jing2024dsbench, bendinelli2025exploring, grosnit2024large}, and broader ML workflows such as data preprocessing, model selection, and hyperparameter tuning~\cite{liu2024large, zhang2023automl, mohr2018ml}.
Prior work in this area can be broadly categorized into agent scaffolding and specialized training approaches. Early scaffolding-based methods structure in-context reasoning and tool use via predefined agent frameworks, but often exhibit unstable prompting gains~\cite{auffarth2023generative, wang2024openhands, jiang2025aide, errica2025did} and susceptibility to local optima~\cite{du2025automlgen}. More recent research has shifted toward LLM-based multi-agent systems~(MAS), which decompose MLE workflows across interacting agents, tools, and memory to improve robustness and performance~\cite{li2024autokaggle, huang2023mlagentbench, fang2025mlzero}. Complementary approaches further incorporate explicit exploration mechanisms~\cite{ou2025automind, yang2025r, du2025automlgen} or learning-based optimization, particularly reinforcement learning, to directly improve agent performance on ML engineering tasks~\cite{yang2025reinforcement}.
Distinct from prior approaches, \method designs an orchestrator that dynamically selects refinement targets and operations rather than coordinating fixed agent roles. Optimized through an online reinforcement learning paradigm, this design enables principled exploration and learning over long-horizon MLE workflows.


\section{Method}
\subsection{Problem Setup: Iterative MLE with Executable Feedback}

We consider iterative machine learning engineering (MLE) tasks where progress is made through
iterative code generation, execution, debugging, and evaluation.
A task instance $x$ provides a natural-language specification, dataset files, a target metric, and a
required submission format. At each round, an executable solution attempt is produced typically as code that generates a submission file and evaluated by an execution environment $\mathcal{E}$.
The environment executes the code, validates the submission format, evaluates the resulting prediction under the task-specific metric, and returns a scalar score.
If execution fails, $\mathcal{E}$ instead returns structured error feedback that can be used for debugging.
The goal is to maximize the best score achieved within a finite budget of solution attempts, i.e., $\max \mathbb{E}[\max_{1 \le t \le T} r_t]$, where $r_t$ denotes the score obtained by the $t$-th solution attempt.

\subsection{\method Framework}
\label{sec:overview}
\method is a hierarchical framework that separates long-horizon decision-making from low-level execution.
It consists of three interacting components: an \emph{Orchestrator}, a set of on-demand \emph{Sub-Agents}, and a
standardized \emph{Tool} interface connected to the execution environment.
The Orchestrator is responsible for deciding what solution direction should be explored or refined next.
Sub-Agents are responsible for implementing and debugging concrete solution attempts.
Tools mediate the interaction between them by constructing Sub-Agent contexts, executing code, collecting
environment feedback, and returning compact summaries and scores.

At initialization, the Orchestrator constructs a root state from the task instance.
This root state captures task-level constraints and initial understandings and serves as the starting point for subsequent refinement.
After initialization, each turn in \method corresponds to one high-level solution refinement attempt. At each turn, the Orchestrator issues a tool call specifying three pieces of information: a previous attempt to refine (if present), detailed new implementation instruction, and optionally a set of previous attempts to use as references. The Tool interface converts this call into a fresh Sub-Agent context, combining the task specification, the selected previous attempt, the Orchestrator instruction, and the tagged references.
The Sub-Agent then implements the requested solution direction through code generation, execution, and
debugging within the environment $\mathcal{E}$.
After the Sub-Agent terminates, the Tool interface returns a compact result to the Orchestrator, including the evaluation score when a valid submission is produced, together with a structured summary.
If execution still fails after the debugging budget is exhausted, the returned result records the failure status and diagnostic summary instead of a valid score.
This result is appended to the Orchestrator history as a standard tool response and becomes available for future refinement decisions.
The process continues until the time budget or maximum number of turns is reached, after which \method
returns the best-scoring solution attempt observed during the interaction.

\subsection{Orchestrator as Decision Layer}
The Orchestrator is the persistent decision-making component of \method. Across turns, it maintains a standard agent context consisting of a sequential record of tool calls and tool responses.
This history gives the Orchestrator a score-annotated view of the solution process, including which
attempts were made, how they performed, and what high-level diagnostics were observed.
Unlike a monolithic agent, the Orchestrator does not directly accumulate raw code, execution logs,
debugging traces, or full Sub-Agent conversations.
It therefore reasons over long-horizon progress at the level of strategic decisions and outcomes.
Each turn, the Orchestrator controls three aspects of the next refinement.
First, it selects previous attempts to be refined, allowing it to exploit strong solutions
or revisit promising but underdeveloped directions.
Second, it writes a concrete implementation instruction for the next Sub-Agent, specifying modeling
choices, preprocessing strategies, hyperparameters, validation protocols, or ensembling plans.
Third, it may tag previous attempts as references, allowing the next Sub-Agent to reuse useful
information from selected prior solutions without exposing the entire execution history.
Thus, at inference time, the Orchestrator dynamically decides what to refine, how to refine it and receives results of the refinement.


\subsection{Tools as Mediation Layer}
Tools define the communication interface between the Orchestrator and Sub-Agents.
They translate high-level refinement decisions from the Orchestrator into executable Sub-Agent
contexts, and translate Sub-Agent execution outcomes back into compact tool responses.
This mediation layer is central to \method: it determines what information is exposed to Sub-Agents
for execution and what information is returned to the Orchestrator for future decisions.

\textbf{Tool Inputs.}
A tool call specifies the refinement attempt to execute.
Concretely, it contains three types of information: the parent attempt to refine, the implementation
instruction generated by the Orchestrator, and optional reference tags pointing to previous attempts.
The parent attempt provides the immediate refinement target.
The instruction specifies what the Sub-Agent should implement.
The reference tags allow the Orchestrator to selectively expose useful prior attempts concisely.

\textbf{Tool Execution.}
Given a tool call, the Tool interface constructs a fresh Sub-Agent context from the provided inputs
and manages the full execution lifecycle of the corresponding Sub-Agent.
It launches the Sub-Agent on the allocated compute resources, mediates its interaction with the environment $\mathcal{E}$, and monitors the process until the Sub-Agent terminates and summarizes the execution.
Multiple Sub-Agents can be launched by multiple, parallel tool calls when parallel exploration is desired, while
their contexts and execution histories remain isolated from one another.

\textbf{Tool Outputs.}
After execution, the Tool interface converts the Sub-Agent trajectory into a compact response for the
Orchestrator.
The response contains the evaluation score if submission is valid and the structured summary.
Thus, Tools enforce the information boundary that allows the Orchestrator to reason over long-horizon
progress without accumulating noisy low-level execution history.

\subsection{Sub-Agents as Execution Layer}
Sub-Agents constitute the execution layer of \method.
While the Orchestrator decides what direction to pursue, each Sub-Agent is responsible for turning one
high-level instruction into an executable solution attempt.
A Sub-Agent is instantiated with a fresh context prepared by the Tool interface, including the task
information needed for execution, the Orchestrator instruction, and any selected references.
It does not inherit the Orchestrator's history or other Sub-Agents' contexts.

Within its local context, the Sub-Agent performs a bounded execution-debugging loop.
It first writes code intended to produce a valid submission file.
The environment $\mathcal{E}$ executes the code and checks whether the submission is valid.
If execution fails or the submission is invalid, the environment returns structured feedback, such as
error messages or validation failures, which the Sub-Agent uses to revise the code.
This process continues until a valid submission is produced or the debugging budget is exhausted.   When a valid submission is produced, $\mathcal{E}$ evaluates it under the task-specific metric and returns a scalar score.
If the Sub-Agent fails to produce a valid submission within the budget, the attempt is marked as
failed and no valid score is assigned.
In both cases, the Sub-Agent summarizes its execution outcome.
The Tool interface returns this summary together with the score when available to the Orchestrator.
This design keeps low-level implementation details local to the Sub-Agent while exposing only
decision-relevant outcomes to the long-horizon decision layer.

\subsection{Iterative Online Optimization} 
We propose an iterative online optimization framework for \method that tightly couples \emph{tree-based trajectory sampling} and \emph{online reinforcement learning}. The framework alternates between structured trajectory collection and policy optimization, where tree-based sampling enables efficient reuse of costly execution feedback and naturally induces contrastive supervision signals for long-horizon decision-making. 
Through iterative data collection, ranking-based labeling, and online contrastive updates, \method jointly improves high-level orchestration and low-level execution, enabling efficient training for complex MLE tasks in expensive execution environments.

\textbf{Training-Time Solution Refinement Tree.}
For each task instance, we first construct an initial root state from the task specification.
Starting from this root, the current Orchestrator policy samples alternative refinement instructions.
Each sampled instruction is executed by a Sub-Agent through the Tool interface, producing a new
solution attempt with a compact summary and score.
The resulting attempt becomes a child node in the refinement tree.
Repeating this process for multiple levels yields a tree whose nodes are score-annotated solution
attempts and whose edges are Orchestrator refinement decisions.
Specifically, we use binary expansion for tractable data collection.
At each non-leaf node, the Orchestrator samples two alternative refinement instructions from the same
parent state.
Both instructions are executed independently, producing two child attempts.

\textbf{Preference Construction.}
The refinement tree is labeled using branch-level downstream returns rather than immediate node
scores.
For an internal node $u$, let $c_1$ and $c_2$ denote the two child branches generated from the same
parent state.
Each child branch is assigned a downstream return, $R(c)=\max_{v \in \mathrm{subtree}(c)} r(v)$, where
$r(v)$ is the score obtained by the solution attempt at node $v$.
The child branch with the higher downstream return is treated as the preferred continuation, while the
other branch is treated as the less preferred continuation.

This construction preserves the long-horizon best-score objective while producing local comparisons
between alternatives that share the same prefix.
Unlike immediate-score comparisons, branch-level returns do not penalize a refinement direction merely
because its first attempt has a lower score but later improves substantially.
Unlike a global winner-take-all labeling scheme, the tree provides multiple common-prefix preference
pairs rather than treating all non-winning trajectories as equally negative.
As a result, the preference signal focuses on the divergent refinement decision at each branching
state while retaining downstream performance as the supervision signal.

\textbf{Online Reinforcement Training.}
At online iteration $m$, the refinement trees are collected using the current Orchestrator policy $
\pi^{(m)}$.
The preference pairs extracted from these trees are then used to train the next Orchestrator policy.
For each branching state, let $s_u$ denote the Orchestrator context at node $u$, and let $\Delta\tau_u^+$ denote the preferred continuation selected by the downstream-return labeling above.
Let $\mathcal{C}(u)$ denote the set of candidate continuations compared at node $u$.
For a trajectory-level score $r(s_u,\Delta\tau)$, the corresponding optimal policy has the energy-based
form 
\begin{equation}\label{eq:ebm_policy}
\pi^\star(\Delta\tau \mid s_u)
=
\frac{1}{Z(s_u)}
\pi^{(m)}(\Delta\tau \mid s_u)
\exp\left(\frac{1}{\beta} r(s_u,\Delta\tau)\right),
\end{equation}
where $Z(s_u)$ is the normalizer and $\beta$ controls the strength of regularization toward the previous policy. We optimize this score with a ranking-based NCE objective that assigns higher probability to the
preferred continuation:
\begin{equation}
\small
\mathcal{L}_{\mathrm{orch}}^{(m+1)}
=
-\mathbb{E}_{u}
\log
\frac{
\exp \left( r_{\theta^{(m+1)}}(s_u,\Delta\tau_u^+) \right)
}{
\sum_{\Delta\tau \in \mathcal{C}(u)}
\exp \left( r_{\theta^{(m+1)}}(s_u,\Delta\tau) \right)
}.
\end{equation}
This energy-based form~\eqref{eq:ebm_policy} implies that, up to a state-dependent normalizing constant, the trajectory score
can be represented by the $\log$-ratio between the updated policy and the reference policy.
Since the constant is shared by all candidates in $\mathcal{C}(u)$ and cancels in the ranking objective, we parameterize the score as
\begin{equation}
r_\theta(s_u,\Delta\tau)
=
\beta
\log
\frac{
\pi_\theta(\Delta\tau \mid s_u)
}{
\pi^{(m)}(\Delta\tau \mid s_u)
}.
\end{equation}


Substituting this score into the ranking objective yields the final Orchestrator loss:
\begin{equation}
\small
\mathcal{L}_{\mathrm{orch}}^{(m+1)}
=
-\mathbb{E}_{u}\bigg[
\beta \log
\frac{
\pi_{\theta^{(m+1)}}(\Delta\tau_u^+ \mid s_u)
}
{
\pi^{(m)}(\Delta\tau_u^+ \mid s_u)
}
- \log\bigg({
\sum_{\Delta\tau \in \mathcal{C}(u)} 
\left(
\frac{
\pi_{\theta^{(m+1)}}(\Delta\tau \mid s_u)
}{
\pi^{(m)}(\Delta\tau \mid s_u)
}\right)^\beta
\bigg)\bigg]
}.
\end{equation}
Here, $\pi^{(m)}$ serves as the reference policy for iteration $m+1$.
This objective optimizes Orchestrator decisions using downstream execution outcomes, while the $KL$-like
$\log$-ratio parameterization keeps each online update anchored to the previous policy.

\textbf{Self-reinforcement for Sub-Agents.}
In addition to optimizing the Orchestrator, we further improve the capability of Sub-Agents through supervised fine-tuning. Specifically, for each refinement tree, we extract Sub-Agent trajectories associated with the highest-scoring refinement paths, and construct instruction-response pairs from the Sub-Agents’ own successful trajectories. These data are used to fine-tune the Sub-Agents, thereby improving their coding, instruction-following ability, and robustness to execution feedback. This self-reinforcement mechanism enables Sub-Agents to improve alongside the Orchestrator, leading to more faithful execution of high-level decisions.

\section{Experiments}




\subsection{Experiment Setups}
\begin{table}[t]
\centering
\small
\setlength{\tabcolsep}{5pt}
\renewcommand{\arraystretch}{1.15}

\caption{HumanRank scores ($\uparrow$ indicates higher is better) across different agent and model configurations. \textit{Dojo Agent} denotes a single-agent setup, while \method denotes the proposed orchestrator--sub-agent framework. "+ self" indicates the same model is used for both Orchestrator and sub-agents. Column \textbf{All} reports the average across the four task categories. All results are reported to four decimal places. \textbf{Bold} indicates the highest score in each category, and \underline{underlining} indicates the second-highest score. See details about mean and variance of scores in Appendix~\ref{app:mean-var-eval}}.
\begin{tabular}{
l l
>{\columncolor{lightHR}}c
>{\columncolor{liteBlue}}c
>{\columncolor{liteGreen}}c
>{\columncolor{litePurple}}c
>{\columncolor{liteOrange}}c
}

\toprule
\textbf{Agent Framework}
& \textbf{Model Configuration}
& \multicolumn{5}{c}{\textbf{HumanRank} ($\uparrow$)} \\
\cmidrule(lr){3-7}
& & \textbf{All} & \textbf{MLE-Lite} & \textbf{Tabular} & \textbf{NLP} & \textbf{CV} \\
\midrule
Dojo Agent
& Qwen3-4B
& 0.1878 & 0.1738 & 0.2045 & 0.1879 & 0.2005 \\
Dojo Agent
& o4-mini
& 0.4832 & 0.5728 & 0.3957 & 0.4704 & 0.3942 \\
\method
& o4-mini + self
& \textbf{0.5465} & \underline{0.6467} & 0.3914 & \textbf{0.5862} & \textbf{0.4555} \\
\method
& Qwen3-4B + o4-mini
& 0.4613 & 0.5990 & 0.3783 & 0.4110 & 0.3004 \\
\method
& Qwen3-4B-SFT + o4-mini
& 0.5061 & 0.6215 & \underline{0.4047} & 0.4407 & 0.4242 \\
\method
& Qwen3-4B-RL + o4-mini
& \underline{0.5360} & \textbf{0.6574} & \textbf{0.4132} & \underline{0.4838} & \underline{0.4506} \\
\midrule
Dojo Agent
& Qwen3-30B-Coder
& 0.3302 & 0.4379 & 0.3047 & 0.2683 & 0.1854 \\
Dojo Agent
& Qwen3-30B-Coder-SFT-Dojo
& 0.3857 & 0.4968 & 0.3426 & 0.3236 & 0.2514  \\
\method
& Qwen3-30B-Coder + self
& 0.3652 & 0.4916 & 0.3008 & 0.3380 & 0.1889 \\
\method
& Qwen3-30B-Coder-SFT + self
& \underline{0.4197} & \underline{0.5365} & \underline{0.3514} & \textbf{0.3866} & \underline{0.2728} \\
\method
& Qwen3-30B-Coder-RL + self
& \textbf{0.4515} & \textbf{0.5897} & \textbf{0.3577} & \underline{0.3722} & \textbf{0.3265} \\
\bottomrule
\end{tabular}
\vspace{-1.5ex}
\label{tab:main}
\end{table}

\textbf{Datasets and Tasks.}
We adopt the comprehensive MLE-Dojo~\cite{qiang2025mledojo} dataset and benchmark, as it covers MLE tasks for both training and evaluation. Specifically, we retain the 150:50 training-evaluation split used by MLE-Dojo. 
MLE-Dojo supports interactive feedback, enabling agents to obtain exact feedback after each interaction. This facilitates refinement of strategies and performance improvement in complex, long-horizon processes.
Both training and evaluation tasks cover a wide range of modalities (\eg, tabular, computer vision, or natural language processing), domains (\eg, finance, medicine, or media), and objectives (\eg, classification, regression, or generation). We also follow the split of evaluation tasks into MLE-Lite, Tabular, CV, and NLP. We adopt \textit{HumanRank} as the evaluation metric, which measures the relative performance by the fraction of human competitors it outperforms on the leaderboard. HumanRank is normalized to $[0,1]$, strictly aligned with the original task-specific metrics, and averaged across public and private leaderboards to ensure fair and comparable evaluation across diverse MLE tasks. See Appendix~\ref{subsec:mledojo} for details about the dataset. 

\textbf{LLMs as Orchestrator and Sub-Agents.} 
We consider a diverse set of cutting-edge large language models (LLMs) within \method, spanning a wide range of model scales and encompassing both open-source and closed-source models. Specifically, we use Qwen3-4B-Instruct-2507~\cite{qwen3technicalreport} and Qwen3-Coder-30B-A3B-Instruct~\cite{qwen3technicalreport} as open-source models at different scales, and o4-mini~\cite{o4-mini} and GPT-5-nano~\cite{gpt-5-nano} as closed-source models of varying sizes.
To comprehensively demonstrate the effectiveness of \method, we conduct extensive experiments across a wide spectrum of model combinations. For convenience of notation, we denote each configuration in the form Model A + Model B, where Model A serves as the Orchestrator and Model B acts as sub-agents.
For open-source models, we adopt the default optimal temperature of 0.7; for closed-source models, we use the default temperature setting of 1.0. We keep all other configurations as default.

\textbf{Agent and Environment Design.}
We refer to the agent scaffold for benchmarking in MLE-Dojo as \textit{Dojo Agent}, and our method as \method. We adapt \textit{Dojo Agent} to make it more capable of handling extended time horizons and increased numbers of interactions with the execution environment. Specifically, we introduce a maximum token limit for each single-round feedback from the environment and apply an effective truncation strategy. The context and maximum output lengths are determined by the the underlying model and are not constrained in either agent, allowing the agents to leverage their full capabilities during long-horizon interactions.
For each task and each run, both agents are allowed up to a maximum of 12 hours of execution time.
For each task and each model configuration, we run three parallel trials and report the best-of-three HumanRank score as the final performance correspondingly. See implementation details of both agents in Appendix~\ref{app:agent-env-design}.


\textbf{Training Strategy.}
For Qwen3-Coder-30B-A3B-Instruct, we train the model to serve both as Orchestrator and Sub-Agents. We first perform supervised fine-tuning (SFT) on 100 high-quality \method trajectories, each comprising one complete Orchestrator trajectory and multiple complete sub-agent trajectories and obtain Qwen3-30B-Coder-SFT. We also perform SFT on 100 high-quality Dojo Agent trajectories, yielding Qwen3-30B-Coder-SFT-Dojo as a monolithic trained baseline. In addition, we fine-tune Qwen3-4B-Instruct-2507 using only Orchestrator trajectories, resulting in Qwen3-4B-SFT.\footnote{We find that Qwen3-4B-Instruct-2507 exhibits significantly weaker performance when used as a Sub-Agent due to model scale and applying SFT to Qwen3-4B as Dojo Agent can even degrade the performance. See Appendix~\ref{app:4bfail} for detailed analyses.} 
We further apply reinforcement learning (RL) to the SFT models. Specifically, we pair Qwen3-4B-SFT as the Orchestrator with o4-mini as Sub-Agents and apply RL training, resulting in Qwen3-4B-RL. We conduct RL training for Qwen3-30B-Coder-SFT to simultaneously serve as both the Orchestrator and Sub-Agents, yielding Qwen3-30B-Coder-RL. 

\subsection{Main Results}
Table~\ref{tab:main} reports HumanRank scores across agent frameworks, model configurations, and evaluation task categories. 
For Qwen3-30B-Coder, \method improves the model from 0.3302 as a monolithic Dojo Agent to 0.3652 when the same model serves as both Orchestrator and Sub-Agent, showing that the hierarchical framework provides gains even without additional training. To further distinguish architecture gains from SFT gains, we compare against Qwen3-30B-Coder-SFT-Dojo, a monolithic Dojo Agent trained with the same SFT protocol. Although SFT improves this trained monolithic baseline, \method with Qwen3-30B-Coder-SFT + self
further reaches 0.4197, indicating that the hierarchical architecture contributes beyond supervised fine-tuning alone. Finally, online RL further improves the \method model to 0.4515, resulting in at most 36.7\% relative performance gain over the Dojo Agent baseline, demonstrating effectiveness of improving Orchestrator decisions and Sub-Agents capabilities with Iterative Online Optimization.

o4-mini improves from 0.4832 as a Dojo Agent to 0.5465 under \method, suggesting that inference-only gains are also evident for frontier closed-source models.
For smaller models, Qwen3-4B performs poorly as a monolithic Dojo Agent, achieving only 0.1878 HumanRank.
However, when served as the Orchestrator within \method with o4-mini Sub-Agents, it reaches 0.4613, and
after SFT and RL further improves to 0.5360.
This nearly matches o4-mini as the Orchestrator under \method, showing that Iterative Online Optimization can make a small model an effective high-level decision layer when paired with stronger execution models.


Overall, these results suggest that models across a wide range of scales can serve as effective Orchestrators within \method. Moreover, additional training on MLE tasks via SFT and RL consistently strengthens their orchestration capability, in some cases enabling smaller models to reach or even surpass the performance of substantially larger models. In contrast, the effectiveness of Sub-Agents is more sensitive to model scale. Small models, such as Qwen3-4B, struggle to reliably follow and execute complex instructions, leading to a noticeable gap between instruction generation as an Orchestrator and instruction execution as a sub-agent, as indicated in Appendix~\ref{app:4bfail}. 

\subsection{Plug-and-Play Capability}
\begin{table*}[t]
\centering
\small
\setlength{\tabcolsep}{6pt}
\renewcommand{\arraystretch}{1.15}
\caption{HumanRank results of Plug-and-Play Capability. Column \textbf{All} reports the average HumanRank across the four task categories.}
\begin{tabular}{
l l
>{\columncolor{lightHR}}c
>{\columncolor{liteBlue}}c
>{\columncolor{liteGreen}}c
>{\columncolor{litePurple}}c
>{\columncolor{liteOrange}}c
}
\toprule
\textbf{Agent Framework}
& \textbf{Model Configuration}
& \multicolumn{5}{c}{\textbf{HumanRank} ($\uparrow$)} \\
\cmidrule(lr){3-7}
& & \textbf{All} & \textbf{MLE-Lite} & \textbf{Tabular} & \textbf{NLP} & \textbf{CV} \\
\midrule
Dojo Agent
& GPT-5-nano
& 0.3946 & 0.5223 & 0.3044 & 0.3139 & 0.2896 \\
\method
& Qwen3-4B-RL +
& \underline{0.4309} & \textbf{0.5679} & \underline{0.3206} & \underline{0.3613} & \underline{0.3162} \\
\method
& Qwen3-30B-Coder-RL +
& \textbf{0.4626} & \underline{0.5594}  & \textbf{0.4406} & \textbf{0.4104} & \textbf{0.3284} \\ 
\bottomrule
\end{tabular}
\vspace{-1.5ex}

\label{tab:plug}
\end{table*}

Within \method, the Orchestrator and Sub-Agents are trained in paired configurations. However, at test time, the trained Orchestrator does not need to be coupled with the exact same Sub-Agents as those used during training. Instead, it generalizes to and consistently improves performance when paired with a broader range of models serving as Sub-Agents. We further study this Plug-and-Play capability of Orchestrators. Table~\ref{tab:plug} reports the performance gains achieved when GPT-5-nano serves as Sub-Agents, with Qwen3-4B-RL and Qwen3-30B-Coder-RL acting as the Orchestrator, respectively. Despite being paired with Sub-Agents that differ from those used during training, the RL-trained Orchestrator consistently yields significant performance improvements against baselines.

\subsection{Ablation Study}
We conduct a series of ablation studies to analyze the contributions of the Orchestrator and the Sub-Agents and to examine the effects of applying reinforcement learning without prior supervised fine-tuning.
To disentangle the sources of performance gains, we further investigate the respective contributions of the Orchestrator and the Sub-Agents to the overall performance gains.
Specifically, we consider two model-combination variants, Qwen3-30B-Coder and Qwen3-30B-Coder-RL, in which the two models alternate as Orchestrator and Sub-Agents. We compare these configurations against the setting in which a single model independently serves as both Orchestrator and Sub-Agent. Table~\ref{tab:self-ablation} shows that employing Qwen3-30B-Coder-RL as either Orchestrator or Sub-Agents consistently improves performance over original model baselines, indicating that both components benefit from training; however, the gains are larger when Qwen3-30B-Coder-RL is used as Orchestrator, highlighting the dominant impact of decision-level optimization.
\begin{table}[h]
\centering
\small
\setlength{\tabcolsep}{-0pt}
\renewcommand{\arraystretch}{1.15}

\begin{minipage}[c]{0.42\linewidth}
\centering
\caption{HumanRank results on Effects of Orchestrator and Sub-Agents.}
\label{tab:self-ablation}
\begin{tabular}{l c}
\toprule
\textbf{Model Configuration} & \textbf{HumanRank (All $\uparrow$)} \\
\midrule
Qwen3-30B-Coder (O)              & 0.3302 \\
Qwen3-30B-Coder-RL + O           & \underline{0.4258} \\
O + Qwen3-30B-Coder-RL           & 0.3826 \\
\midrule
Qwen3-30B-Coder + self           & 0.3652 \\
Qwen3-30B-Coder-RL + self        & \textbf{0.4515} \\
\bottomrule
\end{tabular}
\vspace{-3ex}
\end{minipage}%
\hfill
\begin{minipage}[c]{0.52\linewidth}
\centering
\caption{HumanRank results on Qwen3-30B-Coder-only-RL served as both Orchestrator and Sub-Agents, which denotes RL without SFT. }
\label{tab:onlyrl}
\begin{tabular}{l c}
\toprule
\textbf{Model Configuration} & \textbf{HumanRank (All $\uparrow$)} \\
\midrule
Qwen3-30B-Coder                  & 0.3302 \\
Qwen3-30B-Coder + self           & 0.3652 \\
\midrule
Qwen3-30B-Coder-RL + self        & \textbf{0.4515} \\
Qwen3-30B-Coder-only-RL + self   & \underline{0.3901} \\
\bottomrule
\end{tabular}
\vspace{-3ex}
\end{minipage}

\end{table}


\textbf{RL without SFT.} 
We investigate whether reinforcement learning (RL) alone, without any prior supervised fine-tuning (SFT), can effectively improve model performance within \method. To this end, we directly apply reinforcement learning to Qwen3-30B-Coder, yielding Qwen3-30B-Coder-only-RL, and evaluate its performance under the \method framework. As shown in Table~\ref{tab:onlyrl}, RL without SFT consistently improves performance; however, the overall gains are modest compared to settings that include SFT as warmup initialization.


\subsection{Cumulated Context Length}
\begin{wrapfigure}{r}{0.5\linewidth} 
    \centering
    \vspace{-6pt}
    \includegraphics[width=\linewidth]{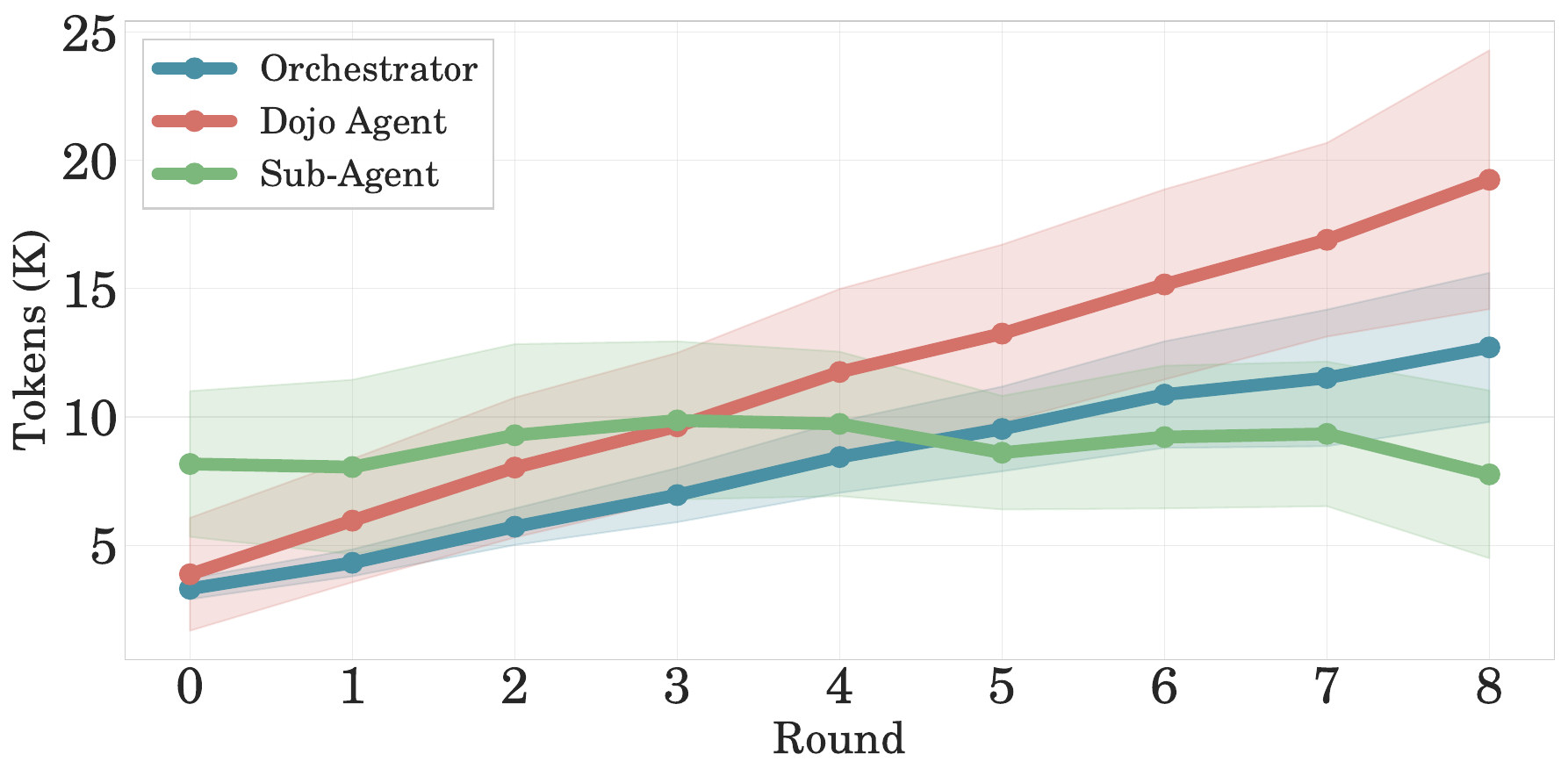}
    \caption{Cumulated Context Length across rounds. X-axis denotes the number of rounds, while y-axis denotes the token count in units of $10^3$. Shaded areas indicate the standard deviation. Contexts of \textit{Dojo Agent} and Orchestrator are averaged over tasks, while contexts of Sub-Agents are averaged per round over tasks.}
    \label{fig:context-length}
    \vspace{-8pt}
\end{wrapfigure}
A core advantage of \method lies in its effective and clean separation of context management between the Orchestrator and the Sub-Agents. As interaction rounds with the environment increase, each sub-agent maintains an independent context with a fixed upper bound that does not grow with the number of interactions. At the same time, although the Orchestrator’s context does grow over time, its growth rate remains modest even as the interaction horizon becomes long. As a result, \method remains robust even under extreme settings with very long time periods and a large number of interactions, whereas the context of the \textit{Dojo Agent} expands more rapidly. Figure~\ref{fig:context-length} illustrates the evolution of context length for the Orchestrator, the Sub-Agents, and the \textit{Dojo Agent} as the number of interaction rounds increases. The figure shows that the Orchestrator’s context length grows at a slower rate than that of the Dojo Agent, while the Sub-Agents’ context lengths remain stable across interaction rounds due to their independence. We further discuss context length scaling in Appendix~\ref{app:context-scale}


\subsection{Orchestrator Instruction Length}
To better understand the insights behind the substantial performance gains after training, we analyze how the length of instructions generated by Orchestrator changes before and after SFT and RL. 
As shown in Appendix~\ref{app:instr-dyna}, Qwen3-4B produces increasingly longer instructions after
SFT and RL, suggesting that a small Orchestrator learns to provide more explicit guidance when
paired with a fixed Sub-Agent.
In contrast, Qwen3-30B-Coder increases instruction length after SFT but remains stable after RL,
indicating that further gains come not merely from longer instructions, but also from improved
Orchestrator--Sub-Agent coordination.
See detailed in Appendix~\ref{app:instr-dyna}.

\vspace{-1ex}
\section{Conclusion}
\vspace{-1ex}
We introduce \method, a hierarchical agent framework that decomposes long-horizon machine learning engineering into coordinated decision-making and execution layers. \method effectively mitigates long-context reasoning challenges and improves scalability under limited computation budgets. We propose a training paradigm for further improving \method.
Extensive experiments across diverse MLE tasks and model configurations demonstrate consistent and substantial performance gains, as well as strong flexibility through plug-and-play orchestration. These results suggest that hierarchical sub-agent unfolding provides a principled and general paradigm for tackling Machine Learning Engineering like complex, long-horizon agentic tasks.

% ...

\bibliography{references}

\begin{thebibliography}{10}

\bibitem{asawa2025train}
P.~Asawa, A.~Zhu, M.~Zaharia, A.~G. Dimakis, and J.~E. Gonzalez.
\newblock How to train your advisor: Steering black-box llms with advisor models.
\newblock {\em arXiv preprint arXiv:2510.02453}, 2025.

\bibitem{auffarth2023generative}
B.~Auffarth.
\newblock Generative ai with langchain.
\newblock {\em Birmingham, UK: Packt Publishing}, 2023.

\bibitem{bendinelli2025exploring}
T.~Bendinelli, A.~Dox, and C.~Holz.
\newblock Exploring llm agents for cleaning tabular machine learning datasets.
\newblock {\em arXiv preprint arXiv:2503.06664}, 2025.

\bibitem{bettini2024benchmarl}
M.~Bettini, A.~Prorok, and V.~Moens.
\newblock Benchmarl: Benchmarking multi-agent reinforcement learning.
\newblock {\em Journal of Machine Learning Research}, 25(217):1--10, 2024.

\bibitem{cemri2025multi}
M.~Cemri, M.~Z. Pan, S.~Yang, L.~A. Agrawal, B.~Chopra, R.~Tiwari, K.~Keutzer, A.~Parameswaran, D.~Klein, K.~Ramchandran, et~al.
\newblock Why do multi-agent llm systems fail?
\newblock {\em arXiv preprint arXiv:2503.13657}, 2025.

\bibitem{chan2024mle}
J.~S. Chan, N.~Chowdhury, O.~Jaffe, J.~Aung, D.~Sherburn, E.~Mays, G.~Starace, K.~Liu, L.~Maksin, T.~Patwardhan, et~al.
\newblock Mle-bench: Evaluating machine learning agents on machine learning engineering.
\newblock {\em arXiv preprint arXiv:2410.07095}, 2024.

\bibitem{chen2025agentflow}
C.~H. Chen and M.~F. Shiu.
\newblock Agentflow: Resilient adaptive cloud-edge framework for multi-agent coordination.
\newblock {\em arXiv preprint arXiv:2505.07603}, 2025.

\bibitem{chen2024autoagents}
G.~Chen, S.~Dong, Y.~Shu, G.~Zhang, J.~Sesay, B.~F. Karlsson, J.~Fu, and Y.~Shi.
\newblock Autoagents: A framework for automatic agent generation, 2024.

\bibitem{chen2023agentverse}
W.~Chen, Y.~Su, J.~Zuo, C.~Yang, C.~Yuan, C.-M. Chan, H.~Yu, Y.~Lu, Y.-H. Hung, C.~Qian, et~al.
\newblock Agentverse: Facilitating multi-agent collaboration and exploring emergent behaviors.
\newblock In {\em The Twelfth International Conference on Learning Representations}, 2023.

\bibitem{chen2025multi}
Y.~Chen, Y.~Wang, S.~Zhu, H.~Yu, T.~Feng, M.~Zhang, M.~Patwary, and J.~You.
\newblock Multi-agent evolve: Llm self-improve through co-evolution.
\newblock {\em arXiv preprint arXiv:2510.23595}, 2025.

\bibitem{comanici2025gemini}
G.~Comanici, E.~Bieber, M.~Schaekermann, I.~Pasupat, N.~Sachdeva, I.~Dhillon, M.~Blistein, O.~Ram, D.~Zhang, E.~Rosen, et~al.
\newblock Gemini 2.5: Pushing the frontier with advanced reasoning, multimodality, long context, and next generation agentic capabilities.
\newblock {\em arXiv preprint arXiv:2507.06261}, 2025.

\bibitem{du2025deepresearch}
M.~Du, B.~Xu, C.~Zhu, X.~Wang, and Z.~Mao.
\newblock Deepresearch bench: A comprehensive benchmark for deep research agents.
\newblock {\em arXiv preprint arXiv:2506.11763}, 2025.

\bibitem{du2025automlgen}
S.~Du, X.~Yan, D.~Jiang, J.~Yuan, Y.~Hu, X.~Li, L.~He, B.~Zhang, and L.~Bai.
\newblock Automlgen: Navigating fine-grained optimization for coding agents.
\newblock {\em arXiv preprint arXiv:2510.08511}, 2025.

\bibitem{errica2025did}
F.~Errica, D.~Sanvito, G.~Siracusano, and R.~Bifulco.
\newblock What did i do wrong? quantifying llms' sensitivity and consistency to prompt engineering.
\newblock In {\em Proceedings of the 2025 Conference of the Nations of the Americas Chapter of the Association for Computational Linguistics: Human Language Technologies (Volume 1: Long Papers)}, pages 1543--1558, 2025.

\bibitem{fang2025mlzero}
H.~Fang, B.~Han, N.~Erickson, X.~Zhang, S.~Zhou, A.~Dagar, J.~Zhang, A.~C. Turkmen, C.~Hu, H.~Rangwala, et~al.
\newblock Mlzero: A multi-agent system for end-to-end machine learning automation.
\newblock {\em arXiv preprint arXiv:2505.13941}, 2025.

\bibitem{grosnit2024large}
A.~Grosnit, A.~Maraval, J.~Doran, G.~Paolo, A.~Thomas, R.~S. H.~N. Beevi, J.~Gonzalez, K.~Khandelwal, I.~Iacobacci, A.~Benechehab, et~al.
\newblock Large language models orchestrating structured reasoning achieve kaggle grandmaster level.
\newblock {\em arXiv preprint arXiv:2411.03562}, 2024.

\bibitem{guo2024ds}
S.~Guo, C.~Deng, Y.~Wen, H.~Chen, Y.~Chang, and J.~Wang.
\newblock Ds-agent: Automated data science by empowering large language models with case-based reasoning.
\newblock {\em arXiv preprint arXiv:2402.17453}, 2024.

\bibitem{hong2024metagpt}
S.~Hong, M.~Zhuge, J.~Chen, X.~Zheng, Y.~Cheng, C.~Zhang, J.~Wang, Z.~Wang, S.~K.~S. Yau, Z.~Lin, L.~Zhou, C.~Ran, L.~Xiao, C.~Wu, and J.~Schmidhuber.
\newblock Metagpt: Meta programming for a multi-agent collaborative framework, 2024.

\bibitem{huang2025r}
C.~Huang, W.~Yu, X.~Wang, H.~Zhang, Z.~Li, R.~Li, J.~Huang, H.~Mi, and D.~Yu.
\newblock R-zero: Self-evolving reasoning llm from zero data.
\newblock {\em arXiv preprint arXiv:2508.05004}, 2025.

\bibitem{huang2023mlagentbench}
Q.~Huang, J.~Vora, P.~Liang, and J.~Leskovec.
\newblock Mlagentbench: Evaluating language agents on machine learning experimentation.
\newblock {\em arXiv preprint arXiv:2310.03302}, 2023.

\bibitem{jain2024livecodebench}
N.~Jain, K.~Han, A.~Gu, W.-D. Li, F.~Yan, T.~Zhang, S.~Wang, A.~Solar-Lezama, K.~Sen, and I.~Stoica.
\newblock Livecodebench: Holistic and contamination free evaluation of large language models for code.
\newblock {\em arXiv preprint arXiv:2403.07974}, 2024.

\bibitem{jiang2025adaptation}
P.~Jiang, J.~Lin, Z.~Shi, Z.~Wang, L.~He, Y.~Wu, M.~Zhong, P.~Song, Q.~Zhang, H.~Wang, et~al.
\newblock Adaptation of agentic ai.
\newblock {\em arXiv preprint arXiv:2512.16301}, 2025.

\bibitem{jiang2025aide}
Z.~Jiang, D.~Schmidt, D.~Srikanth, D.~Xu, I.~Kaplan, D.~Jacenko, and Y.~Wu.
\newblock Aide: Ai-driven exploration in the space of code.
\newblock {\em arXiv preprint arXiv:2502.13138}, 2025.

\bibitem{jimenez2023swe}
C.~E. Jimenez, J.~Yang, A.~Wettig, S.~Yao, K.~Pei, O.~Press, and K.~Narasimhan.
\newblock Swe-bench: Can language models resolve real-world github issues?
\newblock {\em arXiv preprint arXiv:2310.06770}, 2023.

\bibitem{jing2024dsbench}
L.~Jing, Z.~Huang, X.~Wang, W.~Yao, W.~Yu, K.~Ma, H.~Zhang, X.~Du, and D.~Yu.
\newblock Dsbench: How far are data science agents from becoming data science experts?
\newblock {\em arXiv preprint arXiv:2409.07703}, 2024.

\bibitem{li2025matryoshka}
C.~Li, Y.~Zhuang, R.~Qiang, H.~Sun, H.~Dai, C.~Zhang, and B.~Dai.
\newblock Matryoshka pilot: Learning to drive black-box llms with llms.
\newblock In {\em The Thirty-ninth Annual Conference on Neural Information Processing Systems}, 2025.

\bibitem{li2024autokaggle}
Z.~Li, Q.~Zang, D.~Ma, J.~Guo, T.~Zheng, M.~Liu, X.~Niu, Y.~Wang, J.~Yang, J.~Liu, et~al.
\newblock Autokaggle: A multi-agent framework for autonomous data science competitions.
\newblock {\em arXiv preprint arXiv:2410.20424}, 2024.

\bibitem{liang2024encouraging}
T.~Liang, Z.~He, W.~Jiao, X.~Wang, Y.~Wang, R.~Wang, Y.~Yang, S.~Shi, and Z.~Tu.
\newblock Encouraging divergent thinking in large language models through multi-agent debate.
\newblock In {\em Proceedings of the 2024 conference on empirical methods in natural language processing}, pages 17889--17904, 2024.

\bibitem{liu2024large}
S.~Liu, C.~Gao, and Y.~Li.
\newblock Large language model agent for hyper-parameter optimization.
\newblock {\em arXiv preprint arXiv:2402.01881}, 2024.

\bibitem{merrill2026terminal}
M.~A. Merrill, A.~G. Shaw, N.~Carlini, B.~Li, H.~Raj, I.~Bercovich, L.~Shi, J.~Y. Shin, T.~Walshe, E.~K. Buchanan, et~al.
\newblock Terminal-bench: Benchmarking agents on hard, realistic tasks in command line interfaces.
\newblock {\em arXiv preprint arXiv:2601.11868}, 2026.

\bibitem{mohr2018ml}
F.~Mohr, M.~Wever, and E.~H{\"u}llermeier.
\newblock Ml-plan: Automated machine learning via hierarchical planning.
\newblock {\em Machine Learning}, 107(8):1495--1515, 2018.

\bibitem{nam2025mle}
J.~Nam, J.~Yoon, J.~Chen, J.~Shin, S.~{\"O}. Ar{\i}k, and T.~Pfister.
\newblock Mle-star: Machine learning engineering agent via search and targeted refinement.
\newblock {\em arXiv preprint arXiv:2506.15692}, 2025.

\bibitem{gpt-5-nano}
OpenAI.
\newblock Introducing openai gpt-5.
\newblock {\em OpenAI Blog}, 2025.

\bibitem{o4-mini}
OpenAI.
\newblock Introducing openai o3 and o4-mini.
\newblock {\em OpenAI Blog}, 2025.

\bibitem{ou2025automind}
Y.~Ou, Y.~Luo, J.~Zheng, L.~Wei, Z.~Yu, S.~Qiao, J.~Zhang, D.~Zheng, Y.~Mao, Y.~Gao, et~al.
\newblock Automind: Adaptive knowledgeable agent for automated data science.
\newblock {\em arXiv preprint arXiv:2506.10974}, 2025.

\bibitem{ouyang2025kernelbench}
A.~Ouyang, S.~Guo, S.~Arora, A.~L. Zhang, W.~Hu, C.~R{\'e}, and A.~Mirhoseini.
\newblock Kernelbench: Can llms write efficient gpu kernels?
\newblock {\em arXiv preprint arXiv:2502.10517}, 2025.

\bibitem{patwardhan2025gdpval}
T.~Patwardhan, R.~Dias, E.~Proehl, G.~Kim, M.~Wang, O.~Watkins, S.~P. Fishman, M.~Aljubeh, P.~Thacker, L.~Fauconnet, et~al.
\newblock Gdpval: Evaluating ai model performance on real-world economically valuable tasks.
\newblock {\em arXiv preprint arXiv:2510.04374}, 2025.

\bibitem{phan2025humanity}
L.~Phan, A.~Gatti, Z.~Han, N.~Li, J.~Hu, H.~Zhang, C.~B.~C. Zhang, M.~Shaaban, J.~Ling, S.~Shi, et~al.
\newblock Humanity's last exam.
\newblock {\em arXiv preprint arXiv:2501.14249}, 2025.

\bibitem{press2025algotune}
O.~Press, B.~Amos, H.~Zhao, Y.~Wu, S.~K. Ainsworth, D.~Krupke, P.~Kidger, T.~Sajed, B.~Stellato, J.~Park, et~al.
\newblock Algotune: Can language models speed up general-purpose numerical programs?
\newblock {\em arXiv preprint arXiv:2507.15887}, 2025.

\bibitem{qiang2025mledojo}
R.~Qiang, Y.~Zhuang, Y.~Li, R.~Zhang, C.~Li, I.~S.-H. Wong, S.~Yang, P.~Liang, C.~Zhang, B.~Dai, et~al.
\newblock Mle-dojo: Interactive environments for empowering llm agents in machine learning engineering.
\newblock {\em arXiv preprint arXiv:2505.07782}, 2025.

\bibitem{qiang2025mlesmith}
R.~Qiang, Y.~Zhuang, A.~Singh, P.~Liang, C.~Zhang, S.~Yang, and B.~Dai.
\newblock Mle-smith: Scaling mle tasks with automated multi-agent pipeline.
\newblock {\em arXiv preprint arXiv:2510.07307}, 2025.

\bibitem{rank2026posttrainbench}
B.~Rank, H.~Bhatnagar, A.~Prabhu, S.~Eisenberg, K.~Nguyen, M.~Bethge, and M.~Andriushchenko.
\newblock Posttrainbench: Can llm agents automate llm post-training?
\newblock {\em arXiv preprint arXiv:2603.08640}, 2026.

\bibitem{shang2025agentsquare}
Y.~Shang, Y.~Li, K.~Zhao, L.~Ma, J.~Liu, F.~Xu, and Y.~Li.
\newblock Agentsquare: Automatic llm agent search in modular design space, 2025.

\bibitem{starace2025paperbench}
G.~Starace, O.~Jaffe, D.~Sherburn, J.~Aung, J.~S. Chan, L.~Maksin, R.~Dias, E.~Mays, B.~Kinsella, W.~Thompson, et~al.
\newblock Paperbench: Evaluating ai's ability to replicate ai research.
\newblock {\em arXiv preprint arXiv:2504.01848}, 2025.

\bibitem{talebirad2023multi}
Y.~Talebirad and A.~Nadiri.
\newblock Multi-agent collaboration: Harnessing the power of intelligent llm agents.
\newblock {\em arXiv preprint arXiv:2306.03314}, 2023.

\bibitem{tang2023ml}
X.~Tang, Y.~Liu, Z.~Cai, Y.~Shao, J.~Lu, Y.~Zhang, Z.~Deng, H.~Hu, K.~An, R.~Huang, et~al.
\newblock Ml-bench: Evaluating large language models and agents for machine learning tasks on repository-level code.
\newblock {\em arXiv preprint arXiv:2311.09835}, 2023.

\bibitem{qwen3technicalreport}
Q.~Team.
\newblock Qwen3 technical report, 2025.

\bibitem{toledo2025ai}
E.~Toledo, K.~Hambardzumyan, M.~Josifoski, R.~Hazra, N.~Baldwin, A.~Audran-Reiss, M.~Kuchnik, D.~Magka, M.~Jiang, A.~M. Lupidi, et~al.
\newblock Ai research agents for machine learning: Search, exploration, and generalization in mle-bench.
\newblock {\em arXiv preprint arXiv:2507.02554}, 2025.

\bibitem{wan2025rema}
Z.~Wan, Y.~Li, X.~Wen, Y.~Song, H.~Wang, L.~Yang, M.~Schmidt, J.~Wang, W.~Zhang, S.~Hu, and Y.~Wen.
\newblock Rema: Learning to meta-think for llms with multi-agent reinforcement learning, 2025.

\bibitem{wang2024openhands}
X.~Wang, B.~Li, Y.~Song, F.~F. Xu, X.~Tang, M.~Zhuge, J.~Pan, Y.~Song, B.~Li, J.~Singh, et~al.
\newblock Openhands: An open platform for ai software developers as generalist agents.
\newblock {\em arXiv preprint arXiv:2407.16741}, 2024.

\bibitem{wang2024simulating}
Y.~Wang, Y.~Chen, F.~Zhong, L.~Ma, and Y.~Wang.
\newblock Simulating human-like daily activities with desire-driven autonomy.
\newblock {\em arXiv preprint arXiv:2412.06435}, 2024.

\bibitem{wei2025browsecomp}
J.~Wei, Z.~Sun, S.~Papay, S.~McKinney, J.~Han, I.~Fulford, H.~W. Chung, A.~T. Passos, W.~Fedus, and A.~Glaese.
\newblock Browsecomp: A simple yet challenging benchmark for browsing agents.
\newblock {\em arXiv preprint arXiv:2504.12516}, 2025.

\bibitem{wei2025reactplannercentric}
X.~Wei, Y.~Dong, X.~Wang, X.~Zhang, Z.~Zhao, D.~Shen, L.~Xia, and D.~Yin.
\newblock Beyond react: A planner-centric framework for complex tool-augmented llm reasoning, 2025.

\bibitem{xie2024osworld}
T.~Xie, D.~Zhang, J.~Chen, X.~Li, S.~Zhao, R.~Cao, T.~J. Hua, Z.~Cheng, D.~Shin, F.~Lei, et~al.
\newblock Osworld: Benchmarking multimodal agents for open-ended tasks in real computer environments.
\newblock {\em Advances in Neural Information Processing Systems}, 37:52040--52094, 2024.

\bibitem{yan2025selftalk}
B.~Yan, Z.~Zhou, L.~Zhang, L.~Zhang, Z.~Zhou, D.~Miao, Z.~Li, C.~Li, and X.~Zhang.
\newblock Beyond self-talk: A communication-centric survey of llm-based multi-agent systems, 2025.

\bibitem{yang2025reinforcement}
S.~Yang, J.~He-Yueya, and P.~Liang.
\newblock Reinforcement learning for machine learning engineering agents.
\newblock {\em arXiv preprint arXiv:2509.01684}, 2025.

\bibitem{yang2025r}
X.~Yang, X.~Yang, S.~Fang, B.~Xian, Y.~Li, J.~Wang, M.~Xu, H.~Pan, X.~Hong, W.~Liu, et~al.
\newblock R\&d-agent: Automating data-driven ai solution building through llm-powered automated research, development, and evolution.
\newblock {\em arXiv preprint arXiv:2505.14738}, 2025.

\bibitem{yao2024tau}
S.~Yao, N.~Shinn, P.~Razavi, and K.~Narasimhan.
\newblock $\tau$-bench: A benchmark for tool-agent-user interaction in real-world domains.
\newblock {\em arXiv preprint arXiv:2406.12045}, 2024.

\bibitem{zhang2023automl}
S.~Zhang, C.~Gong, L.~Wu, X.~Liu, and M.~Zhou.
\newblock Automl-gpt: Automatic machine learning with gpt.
\newblock {\em arXiv preprint arXiv:2305.02499}, 2023.

\bibitem{zhang2026agentorchestra}
W.~Zhang, L.~Zeng, Y.~Xiao, Y.~Li, C.~Cui, Y.~Zhao, R.~Hu, Y.~Liu, Y.~Zhou, and B.~An.
\newblock Agentorchestra: Orchestrating multi-agent intelligence with the tool-environment-agent(tea) protocol, 2026.

\bibitem{zhang2024chainagents}
Y.~Zhang, R.~Sun, Y.~Chen, T.~Pfister, R.~Zhang, and S.~{\"O}. Arik.
\newblock Chain of agents: Large language models collaborating on long-context tasks, 2024.

\bibitem{zhang2025lazyagent}
Z.~Zhang, X.~Li, Y.~Lin, H.~Liu, R.~Chandradevan, L.~Wu, M.~Lin, F.~Wang, X.~Tang, Q.~He, and S.~Wang.
\newblock Unlocking the power of multi-agent llm for reasoning: From lazy agents to deliberation, 2025.

\bibitem{zhou2023webarena}
S.~Zhou, F.~F. Xu, H.~Zhu, X.~Zhou, R.~Lo, A.~Sridhar, X.~Cheng, T.~Ou, Y.~Bisk, D.~Fried, et~al.
\newblock Webarena: A realistic web environment for building autonomous agents.
\newblock {\em arXiv preprint arXiv:2307.13854}, 2023.

\end{thebibliography}
\bibliographystyle{abbrv}


\clearpage
\appendix

\section{Benchmark and Metric Details}
\label{subsec:mledojo}

\textbf{MLE-Dojo Benchmark.}
We conduct all experiments on \textbf{MLE-Dojo}~\cite{qiang2025mledojo}, a large-scale benchmark designed to evaluate autonomous agents on realistic machine learning engineering (MLE) tasks.
MLE-Dojo consists of 200 end-to-end Kaggle-style MLE tasks, including 150 training tasks and 50 held-out evaluation tasks, following the official benchmark split.
Each task requires agents to perform a complete MLE workflow, including data preprocessing, feature engineering, model selection, training, evaluation, and submission.

A key feature of MLE-Dojo is its interactive environment, where agents can iteratively submit solutions to a leaderboard and receive exact evaluation feedback after each interaction.
This enables agents to refine strategies over long horizons, closely resembling real-world iterative MLE pipelines.

For evaluation, tasks are grouped into four categories following MLE-Dojo:
\textit{MLE-Lite} (21 tasks), \textit{Tabular} (10 tasks), \textit{CV} (9 tasks), and \textit{NLP} (10 tasks), corresponding to lightweight benchmark tasks, structured tabular problems, computer vision tasks, and natural language processing tasks, respectively.

\textbf{HumanRank Metric.}
We adopt \textbf{HumanRank} as the primary evaluation metric.
HumanRank measures the relative ranking of an agent submission against human competitors on the leaderboard.
Suppose a submission ranks at position $p$ among $N$ total submissions; the HumanRank score is defined as:
\begin{equation}
s = 1 - \frac{p}{N}.
\end{equation}
Thus, HumanRank reflects the fraction of human participants that the agent outperforms, with higher values indicating stronger performance.

HumanRank is a normalized metric within the range $[0,1]$, strictly aligned with the original task-specific leaderboard metrics (e.g., accuracy, RMSE, or BLEU).
To mitigate potential bias between public and private leaderboards, HumanRank is computed independently on both leaderboards and averaged as the final reward score.
This normalization enables fair and comparable evaluation across heterogeneous MLE tasks with different scoring scales.


\section{Agent and Environment Design}
\label{app:agent-env-design}
\textbf{Dojo Agent.}
The max time limit is set to 12 hours, with agents having full access to their interaction histories. A unified, concise prompt with clear instructions is provided at the beginning, without
additional extraneous information. The maximum input token length is not limited, while each output round is also not limited. We do not restrict the number of submission attempts to enable continuous improvement.
To generate valid submission files and scores, agents must explicitly use the "execute\_code"
command; submissions are not automated. We pre-install commonly used Python packages, though
agents may install additional packages within their generated code as needed.
We make \textbf{several key modifications} to the original MLE-Dojo Agent.
First, we remove the explicit cap on the model’s context length. While MLE-Dojo typically restricts the context window to 50k tokens, we instead allow each model to utilize its native maximum context capacity.
Second, we apply output truncation to environment feedback: when the environment output exceeds 5k tokens, we retain the first 2.5k and the last 2.5k tokens and concatenate them.
Together, these changes enable the Dojo Agent to operate effectively over substantially longer horizons, alleviating premature context saturation while preserving critical execution signals from the environment.

\textbf{\method.}
The max time limit is set to 12 hours.  A unified, concise prompt with clear instructions is provided at the beginning, without
additional extraneous information. 
As with the Dojo Agent, we do not impose an explicit limit on the model’s maximum context length, allowing it to fully utilize its native context capacity. We also adopt the same environment output truncation mechanism: when the environment feedback exceeds 5k tokens, the first and last 2.5k tokens are retained and concatenated.
Each Sub-Agent is allowed to perform up to 10 debugging iterations within a round. If this limit is exceeded, the corresponding run is terminated and the resulting score is recorded as N/A.

\section{SFT Data Collection}
\label{app:data-col}
We collect \method supervised fine-tuning (SFT) data by running the full \method pipeline
with gemini-2.5-flash~\cite{comanici2025gemini} serving simultaneously as both the Orchestrator and
the Sub-Agents.
We also collect Dojo Agent supervised fine-tuning (SFT) data by running Dojo Agent with the same model.
Specifically, we randomly sample 100 training tasks from the MLE-Dojo benchmark
and execute the agentic system on each task with a maximum wall-clock budget of
12 hours.
For each task, execution is early-stopped once a perfect score is achieved.

This procedure yields 100 complete agentic trajectories, one per task.
Each complete trajectory consists of a single Orchestrator trajectory spanning
multiple rounds, together with multiple corresponding Sub-Agent execution
trajectories instantiated across rounds.
We retain the 100 Orchestrator trajectories as SFT data to fine-tune
Qwen3-4B-Instruct-2507.
In addition, we use the same 100 Orchestrator trajectories together with 300
selected Sub-Agent trajectories to separately fine-tune
Qwen3-Coder-30B-A3B-Instruct, resulting in SFT-initialized versions of
both the Orchestrator and Sub-Agent models.

\section{Mean and Variance of Evaluation}
\label{app:mean-var-eval}

For each of the 50 evaluation tasks, we run three independent trials and report the per-task best
score
before averaging across tasks.
This best-of-three protocol follows MLE-Dojo and is designed to reduce the effect of random
execution
failures, such as malformed submissions produced in otherwise reasonable solution attempts.
It also reflects the practical setting of MLE competitions, where multiple submissions are
attempted and
the best valid result is selected.

To address variance concerns, we additionally report the per-task mean score averaged across
tasks, which
is a stricter aggregation metric.
As shown in Table~\ref{tab:mean-var-eval}, the relative ordering of methods is preserved under
this
mean-based evaluation.
The gap between best and mean scores is also consistent across methods, and the median per-task
standard
deviation across the 50 evaluation tasks is approximately $0.07$.

\begin{table}[h]
\centering
\small
\setlength{\tabcolsep}{6pt}
\renewcommand{\arraystretch}{1.15}
\caption{Comparison between per-task best-of-three and per-task mean aggregation across three
independent trials.}
\label{tab:mean-var-eval}
\begin{tabular}{lcc}
\toprule
\textbf{Model Configuration} & \textbf{Avg. Per-Task Best} & \textbf{Avg. Per-Task Mean} \\
\midrule
Dojo Agent: o4-mini & 0.4832 & 0.4053 \\
\method: o4-mini + self & 0.5465 & 0.4707 \\
\method: Qwen3-30B-Coder-RL + self & 0.4515 & 0.3748 \\
\method: Qwen3-4B-RL + o4-mini & 0.5360 & 0.4596 \\
\bottomrule
\end{tabular}
\end{table}
\section{Failure Analysis of Qwen3-4B}
\label{app:4bfail}
In this section, we analyze why we do not employ Qwen3-4B as both the Orchestrator and Sub-Agent in our main experiments. We present representative execution trajectories for the pretrained Qwen3-4B model (Table~\ref{tab:qwen3-4b-all-pretrain-case}) and its SFT-enhanced counterpart (Table~\ref{tab:qwen3-4b-all-sft-case}) on the \texttt{leaf-classification} task from the MLE-Dojo evaluation set.

\textbf{Qwen3-4B: Insufficient Instruction Granularity.}
As shown in Table~\ref{tab:qwen3-4b-all-pretrain-case}, the pretrained Qwen3-4B Orchestrator issues relatively coarse and high-level instructions, such as performing PCA-based feature engineering and training a LightGBM classifier. However, these instructions lack concrete procedural scaffolding (e.g., explicit data schema handling, feature concatenation strategies, and submission formatting constraints). As a result, the Sub-Agent fails to correctly execute the pipeline, exhibiting syntax-level hallucinations and repeated schema mismatches (e.g., missing \texttt{margin\_1} columns). Eventually, the Sub-Agent collapses to a degenerate fallback strategy that outputs uniform class probabilities, yielding near-random leaderboard performance (Position 1546/1595). This indicates that the pretrained Orchestrator does not provide sufficient operational detail for reliable downstream execution.

\textbf{Qwen3-4B-SFT: Improved Instructions but Weak Instruction Following Ability.}
After supervised fine-tuning, the Qwen3-4B-SFT Orchestrator produces substantially more detailed and structured instructions (Table~\ref{tab:qwen3-4b-all-sft-case}), including explicit data loading steps, feature concatenation procedures, LightGBM hyperparameters, and submission constraints. This demonstrates that SFT successfully transfers procedural instruction templates from the teacher model. However, despite the improved instruction quality, the Qwen3-4B-SFT Sub-Agent still fails to follow these instructions effectively. Concretely, the Sub-Agent repeatedly encounters execution errors such as missing feature columns and fragmented DataFrame insertions, and ultimately fails to produce a valid submission. This highlights that, at this model scale, instruction following and execution grounding remain weak even when the Orchestrator provides detailed guidance.


\section{Analysis of Context Length Scaling}
\label{app:context-scale}
We analyze the empirical growth rates of context length across interaction rounds to characterize the scaling behavior of the Orchestrator \& Sub-Agent framework. As shown in Figure~\ref{fig:context-length}, the Orchestrator exhibits a moderate linear growth in context length, with an estimated slope of \textbf{1201.93 tokens per round}, while the Dojo Agent grows substantially faster, with a slope of \textbf{1870.29 tokens per round}. In contrast, the Sub-Agents maintain an approximately constant context length across rounds, due to their independent and resettable execution contexts.

We further estimate the maximum number of interaction rounds that can be supported under a fixed context budget. Assuming a base model such as Qwen3-30B-Coder~\cite{qwen3technicalreport} with a natural maximum context length of 256K tokens, the Orchestrator in our framework can support approximately $\lfloor 256{,}000 / 1201.93 \rfloor \approx 213$ rounds before exceeding the context limit, whereas the Dojo Agent can support only $\lfloor 256{,}000 / 1870.29 \rfloor \approx 137$ rounds. This indicates that the modular Orchestrator \& Sub-Agent design significantly improves long-horizon scalability by decoupling execution memory from orchestration memory, enabling substantially longer interaction horizons under the same context budget.





\section{Training Strategy and Details}
\textbf{Hardware and Software.}
For the training of both Qwen3-4B-Instruct-2507 and Qwen3-Coder-30B-A3B-Instruct, we conducted all experiments on servers equipped with Intel Xeon Platinum 8580 CPUs and NVIDIA H100 80GB HBM3 GPUs, using Python 3.10.12.


\textbf{Supervised Fine-Tuning.}
For both models, we first perform SFT on 100 high-quality trajectories. Qwen3-Coder-30B-A3B-Instruct is fine-tuned on both the Orchestrator and sub-agents, whereas Qwen3-4B-Instruct-2507 is fine-tuned only on the Orchestrator. Training is conducted on 8 H100 80GB GPUs with the hyperparameters listed in Table~\ref{tab:sft_hyperparams}. 

\textbf{Online Reinforcement Learning.}
For both models, we further perform online RL training on the Orchestrator using a ranking-based NCE objective with trajectory-level feedback. 
Training is conducted on 8 H100 80GB GPUs with the hyperparameters listed in Table~\ref{tab:infonce_hyperparams}.
\begin{table}[t]
\centering
\caption{Hyperparameters for SFT training of MLE agents.}
\label{tab:sft_hyperparams}
\begin{tabular}{lcc}
\toprule
\textbf{Hyperparameter} &
\textbf{Qwen3-4B-Instruct-2507} &
\textbf{Qwen3-Coder-30B-A3B-Instruct} \\
\midrule

max\_length & 32768 & 32768 \\
train\_batch\_size & 4 & 2 \\
total\_epochs & 2 & 2 \\
nnodes & 1 & 1 \\
n\_gpus\_per\_node & 4 & 8 \\

\midrule
tensor\_model\_parallel\_size & 2 & 4 \\
pipeline\_model\_parallel\_size & 1 & 1 \\
context\_parallel\_size & 2 & 2 \\
sequence\_parallel & True & True \\
expert\_model\_parallel\_size & 1 & 8 \\
expert\_tensor\_parallel\_size & 1 & 1 \\
gradient\_checkpointing & True & True \\
dynamic\_batch\_size & True & True \\

\midrule
optimizer & Adam & Adam \\
learning\_rate & $1\times 10^{-5}$ & $1\times 10^{-5}$ \\
lr\_decay\_style & cosine & cosine \\
min\_lr & $1\times 10^{-6}$ & $1\times 10^{-6}$ \\
warmup\_fraction & 0.1 & 0.1 \\
weight\_decay & 0.1 & 0.1 \\
adam\_beta1 & 0.9 & 0.9 \\
adam\_beta2 & 0.95 & 0.95 \\
optimizer\_cpu\_offload & False & True \\

\bottomrule
\end{tabular}
\end{table}

\begin{table}[t]
\centering
\caption{Hyperparameters for Online RL training of MLE agents.}
\label{tab:infonce_hyperparams}
\begin{tabular}{lcc}
\toprule
\textbf{Hyperparameter} &
\textbf{Qwen3-4B-SFT} &
\textbf{Qwen3-Coder-30B-SFT} \\
\midrule

max\_length & 32768 & 32768 \\
train\_batch\_size & 1 & 1 \\
gradient\_accumulation\_steps & 8 & 8 \\
total\_epochs & 2 & 2 \\
nnodes & 1 & 1 \\
n\_gpus\_per\_node & 4 & 8 \\

\midrule
tensor\_model\_parallel\_size & 2 & 1 \\
pipeline\_model\_parallel\_size & 2 & 4 \\
context\_parallel\_size & 1 & 1 \\
expert\_model\_parallel\_size & 1 & 2 \\
expert\_tensor\_parallel\_size & 1 & 1 \\
gradient\_checkpointing & True & True \\
dynamic\_batch\_size & False & False \\

\midrule
optimizer & Adam & Adam \\
learning\_rate & $1\times 10^{-5}$ & $1\times 10^{-5}$ \\
lr\_decay\_style & cosine & cosine \\
warmup\_fraction & 0.1 & 0.1 \\


\bottomrule
\end{tabular}
\end{table}

\section{Case Study on Instruction Dynamics}
\label{app:instr-dyna}
In this section, we conduct qualitative case studies to analyze how Orchestrator instruction length evolves across training stages and to uncover the mechanisms underlying the divergent instruction dynamics between Qwen3-4B and Qwen3-Coder-30B.

Specifically, we measure the average instruction length on evaluation tasks when Qwen3-30B-Coder serves as both Orchestrator and Sub-Agents, as well as when Qwen3-4B serves as Orchestrator with o4-mini as Sub-Agents. 
Table~\ref{tab:instruction-dyna} shows that for Qwen3-4B, the instruction length increases substantially after SFT and RL, and longer instructions correlate with better performance. In contrast, for Qwen3-Coder-30B, the instruction length grows notably after SFT but remains roughly unchanged after RL. This discrepancy can be partly attributed to the fact that during RL, Qwen3-Coder-30B is also trained as Sub-Agents, leading to improved instruction-following ability. As a result, it relies less on longer or more detailed instructions compared to Qwen3-4B, which is trained only as an Orchestrator. We further conduct a case study on the instructions generated by the Orchestrator to gain deeper insights into this phenomenon.
\begin{table}[h]
\centering
\small
\setlength{\tabcolsep}{6pt}
\renewcommand{\arraystretch}{1.2}
\caption{Average instruction token count of Orchestrator over evaluation tasks across different training stages.}
\begin{tabular}{l c c c}
\toprule
\textbf{Model Configuration} & \textbf{Pretrain} & \textbf{SFT} & \textbf{RL} \\
\midrule
Qwen3-Coder-30B + self & 271.62 & 639.10 & 655.63 \\
Qwen3-4B + o4-mini     & 379.31 & 750.06 & 1010.22 \\
\bottomrule
\end{tabular}
\label{tab:instruction-dyna}
\end{table}

\textbf{Instruction Dynamics of Qwen3-4B.}
We first analyze the task \texttt{dog-breed-identification} from the MLE-Dojo evaluation set to study instruction evolution for Qwen3-4B (Table~\ref{tab:instruction-dynamics-qwen3-4b}). 
Prior to training, the pretrained Orchestrator produces relatively concise but technically oriented instructions, primarily focusing on high-level strategies such as class-imbalance-aware data augmentation and loss reweighting. 
After supervised fine-tuning (SFT), the Orchestrator generates substantially longer and more structured instructions, explicitly detailing data loading pipelines, preprocessing steps, model architecture choices (e.g., EfficientNet backbones), training hyperparameters, and inference procedures, indicating the transfer of detailed instructional style and decomposition patterns from the teacher model.

Notably, instruction length continues to increase after reinforcement learning (RL) for Qwen3-4B. This behavior can be attributed to the fact that in this setting, only the Orchestrator is trained during RL while the Sub-Agent remains fixed. Consequently, the Orchestrator is forced to compensate by producing increasingly detailed and explicit instructions to better steer the fixed Sub-Agent toward high-quality solutions. For example, after RL, the Orchestrator introduces finer-grained architectural modifications (e.g., upgrading to EfficientNetB4 with a custom classification head), training heuristics, and system-level considerations, reflecting an emergent strategy of instruction amplification to improve downstream execution.

\textbf{Instruction Dynamics of Qwen3-Coder-30B.}
We next examine one common competition task \texttt{conways-reverse-game-of-life-2020} from the MLE-Dojo evaluation set to analyze instruction evolution for Qwen3-Coder-30B (Table~\ref{tab:instruction-dynamics-qwen3-coder-30b}). 
Before training, the pretrained Orchestrator produces concise and generic task descriptions that mainly specify the high-level objective (e.g., implementing a CNN-based model to reverse Conway’s Game of Life). 
After SFT, the Orchestrator learns to mimic the teacher model’s detailed instructional style, resulting in significantly longer and more structured instructions that explicitly specify data loading, model architecture, training procedures, inference steps, and implementation constraints, suggesting the transfer of instructional verbosity and structured decomposition patterns.

In contrast to Qwen3-4B, instruction length for Qwen3-Coder-30B remains largely unchanged after RL, despite substantial performance improvements. This phenomenon can be explained by the \textit{co-evolution} of the Orchestrator and Sub-Agent during RL. Since the Sub-Agent is simultaneously trained to better follow instructions, the Orchestrator no longer needs to increase instruction verbosity to achieve stronger downstream execution. Instead, it maintains a similar instruction length while relying on the improved instruction-following capability of the Sub-Agent to execute tasks more effectively, highlighting a different coordination strategy from Qwen3-4B.

\newpage
{
\small
\setlength{\LTleft}{0pt}
\setlength{\LTright}{0pt}
\renewcommand{\arraystretch}{1.4}

\begin{longtable}{>{\RaggedRight}p{3.5cm} >{\RaggedRight}p{10.0cm}}

  \caption{Qwen3-4B as Orchestrator and Sub-Agent on \texttt{leaf-classification}} 
  \label{tab:qwen3-4b-all-pretrain-case} \\
  \toprule
  \textbf{Role} & \textbf{Corresponding Trajectory} \\
  \midrule
  \endfirsthead

  \toprule
  \textbf{Role} & \textbf{Corresponding Trajectory} \\
  \midrule
  \endhead

  \midrule
  \multicolumn{2}{r@{}}{\textit{Continued on next page}} \\
  \endfoot

  \bottomrule
  \endlastfoot

  \textit{Qwen3-4B as Orchestrator LLM} &
Perform feature engineering on the margin and texture features using PCA for dimensionality reduction and feature selection via mutual information.

Combine the reduced features with the shape features to create a more compact and informative feature set. 

Apply PCA to margin and texture features separately, retaining only the top 30 principal components to reduce noise and redundancy. 

Use mutual information to select the top 20 features from the combined set. 

Train a LightGBM model on the engineered features with class weights to handle imbalance. 

Evaluate using logarithmic loss and generate a submission file in the correct format.

Focus on improving model performance by capturing subtle patterns in the data that may be missed by raw feature usage.
  \\ \midrule

  \textit{Qwen3-4B as Sub-Agent} &
  \begin{minipage}[t]{\linewidth}
\setlength{\parskip}{0.4em}

\textbf{Attempt 1:} Execute PCA-based feature engineering and LightGBM training. \par
\textbf{Feedback 1:} 
SyntaxError: invalid imaginary literal.

\textbf{Attempt 2:} Retry pipeline with engineered features and probability prediction. \par
\textbf{Feedback 2:} Execution failed due to data schema mismatch and repeated parsing errors.

\textbf{Attempt 3:} Fallback to generate a submission with uniform class probabilities. \par
\textbf{Feedback 3:} Submission generated successfully. 

\textbf{Evaluation Results: }

Position: \textit{1546 / 1595}, Raw Score: \textit{4.595}.
\end{minipage}
\\

\end{longtable}
}

\newpage
{
\small
\setlength{\LTleft}{0pt}
\setlength{\LTright}{0pt}
\renewcommand{\arraystretch}{1.4}

\begin{longtable}{>{\RaggedRight}p{3.5cm} >{\RaggedRight}p{10.0cm}}

  \caption{Qwen3-4B-SFT as both Orchestrator and Sub-Agent} 
  \label{tab:qwen3-4b-all-sft-case} \\
  \toprule
  \textbf{Role} & \textbf{Corresponding Trajectory} \\
  \midrule
  \endfirsthead

  \toprule
  \textbf{Role} & \textbf{Corresponding Trajectory} \\
  \midrule
  \endhead

  \midrule
  \multicolumn{2}{r@{}}{\textit{Continued on next page}} \\
  \endfoot

  \bottomrule
  \endlastfoot

  \textit{Qwen3-4B-SFT as Orchestrator LLM} &
\begin{minipage}[t]{\linewidth}
\setlength{\parskip}{0.4em}

You are a Kaggle grandmaster. Your task is to develop a multi-class classification model to identify 99 plant species using the provided pre-extracted features: shape, margin, and texture. Your solution should leverage a LightGBM classifier.

\textbf{Key Implementation Details:}

1.\ Data Loading: \par
Load \texttt{train.csv}, \texttt{test.csv}, and \texttt{sample\_submission.csv} from the specified \texttt{data\_path}. Combine training and testing data for consistent preprocessing.

2.\ Feature Engineering: \par
The features are provided as individual columns (e.g., \texttt{margin\_1}, \texttt{margin\_2}, \ldots, \texttt{margin\_64}). Concatenate these 64 attributes for each feature type (shape, margin, texture) to form a single feature vector for each image. For example, for the \emph{margin} feature, combine \texttt{margin\_1} through \texttt{margin\_64} into a single column named \texttt{margin\_all}. Do the same for \texttt{shape\_all} and \texttt{texture\_all}. Drop the \texttt{id} column from the combined training data and use it only for the final submission.

3.\ Target Variable: \par
The target variable in the training data is the \texttt{species} column.

4.\ Model Training: \par
Use a LightGBM Classifier (\texttt{LGBMClassifier}) with \texttt{objective='multiclass'}, \texttt{num\_class=99}, and \texttt{metric='multi\_logloss'}. Set \texttt{is\_unbalance=True}, \texttt{n\_estimators=1000}, \texttt{learning\_rate=0.05}, \texttt{min\_child\_samples=20}, \texttt{subsample=0.7}, \texttt{colsample\_bytree=0.7}, \texttt{reg\_alpha=0.1}, \texttt{reg\_lambda=0.1}, \texttt{random\_state=42}, and \texttt{n\_jobs=-1}.

5.\ Prediction: \par
Generate probability predictions for the preprocessed test set and ensure probabilities are within the [0, 1] range.

6.\ Submission File Generation: \par
Create a submission DataFrame with the \texttt{id} column and species probability columns ordered exactly as in \texttt{sample\_submission.csv}. Save \texttt{submission.csv} to the \texttt{output\_path} using absolute paths.

\textbf{Constraints and Guidelines:} \par
Do not use any image data. Ensure no data leakage, preserve test IDs, normalize probabilities if needed, and minimize multi-class log loss.

Your code should be a complete and runnable Python script that performs all necessary steps from loading data to generating the final submission file.

\end{minipage}
  \\ \midrule

  \textit{Qwen3-4B-SFT as Sub-Agent} &
  \begin{minipage}[t]{\linewidth}
  \setlength{\parskip}{0.45em}

  \textbf{Attempt 1:} Implement feature concatenation as instructed and train \texttt{LGBMClassifier}. \par
  \textbf{Feedback 1:} Execution failed:
  KeyError:\ 'margin\_1'.

  \textbf{Attempt 2:} Try to patch schema by inserting / filling missing columns \par
  \textbf{Feedback 2:} Execution failed:\ 
  PerformanceWarning:\ DataFrame is highly fragmented; Consider joining all columns at once using \texttt{pd.concat(axis=1)} \ldots

  \textbf{Final Status:} \textbf{FAILED} (no valid submission / no leaderboard evaluation). \par
  \textbf{Evaluation Results:}\ 
  
  Position:\ \textit{N/A}, Raw Score: \textit{N/A}.
  \end{minipage}
\\

\end{longtable}
}

\newpage
{
\small
\setlength{\LTleft}{0pt}
\setlength{\LTright}{0pt}
\renewcommand{\arraystretch}{1.4}

\begin{longtable}{>{\RaggedRight}p{3.5cm} l >{\RaggedRight}p{7.5cm}}

  \caption{Instruction Dynamics of Qwen3-4B on \texttt{dog-breed-identification}} \label{tab:instruction-dynamics-qwen3-4b} \\
  \toprule
  \endfirsthead

  \toprule
  \endhead

  \midrule
  \multicolumn{3}{r@{}}{\textit{Continued on next page}} \\
  \endfoot

  \bottomrule
  \endlastfoot

  \multicolumn{3}{c}{\textit{Qwen3\_4B\_Pretrain (score = 0.3391)}} \\ \midrule
  \multicolumn{3}{>{\RaggedRight}p{\linewidth}}{
  Explore a novel data augmentation strategy using mixup and cutout with adaptive parameters based on class frequency.
  For rare breeds (fewer than 50 images), apply higher mixup ratios (0.8--1.0) and larger cutout masks (10--20\% of image area) to improve generalization.
  For common breeds (more than 100 images), use lower mixup ratios (0.3--0.5) and smaller cutout masks (5--10\%) to preserve image integrity.
  Implement a dynamic augmentation pipeline that adjusts these parameters during training based on the current batch's class distribution.
  Additionally, incorporate focal loss with $\gamma=2$ to emphasize learning from underrepresented breeds.
  This method balances data augmentation with class imbalance, directly addressing the core challenges of fine-grained dog breed identification.
  The implementation should include proper data loading, augmentation, training loop, and validation metrics to ensure robust performance.
  } \\
  \midrule

  \multicolumn{3}{c}{\textit{Qwen3\_4B\_SFT (score = 0.3828)}} \\ \midrule
  \multicolumn{3}{>{\RaggedRight}p{\linewidth}}{
  Implement a baseline solution using a pre-trained EfficientNetB0 model from the \texttt{timm} library for fine-grained image classification of 120 dog breeds.

  \textbf{Implementation Details:}
  
  (1) Data Loading and Preprocessing: Load train.zip and test.zip using the zipfile module and extract to temporary directories. Create ImageFolder datasets and map filenames to breed labels from labels.csv. Apply standard transforms (resize to 224$\times$224, center crop, normalization). Include random horizontal flip for training.  
  
  (2) Model Architecture: Load a pre-trained EfficientNetB0 model and modify the final classification layer to output 120 classes.  
  
  (3) Training Strategy: Use Adam optimizer and CrossEntropyLoss. Train for 10--15 epochs with batch size 32 or 64. Monitor validation loss and accuracy and save the best checkpoint.  
  
  (4) Inference and Submission: Load the best model, perform inference on test images, generate probability predictions, and format them into submission.csv with column names matching labels.csv.  

  \textbf{Constraints and Best Practices:}
  
  Ensure proper imports, CUDA support, self-contained code, careful memory usage, absolute paths, and strict adherence to sample submission format.
  } \\
  \midrule

  \multicolumn{3}{c}{\textit{Qwen3\_4B\_RL (score = 0.4297)}} \\ \midrule
  \multicolumn{3}{>{\RaggedRight}p{\linewidth}}{
  The goal is to improve dog breed classification accuracy by replacing ResNet50 with EfficientNetB4 and adding a custom classification head.

  The classification head consists of:
  
  (1) Global average pooling,  
  
  (2) A dense layer with 512 units and ReLU activation,  
  
  (3) A final dense layer with 120 units and sigmoid activation.

  \textbf{Key Considerations:}
  
  (1) Framework: TensorFlow/Keras.  
  
  (2) Pre-trained Weights: Load ImageNet-pretrained EfficientNetB4.  
  
  (3) Input Size: Ensure compatibility with EfficientNetB4 recommended resolution.  
  
  (4) Transfer Learning: Fine-tune the entire model.  
  
  (5) Data Loading and Preprocessing: Load train/test datasets, normalize images, and apply extensive data augmentation (flips, rotations, color jitter, random resized crop).  
  
  (6) Loss: SparseCategoricalCrossentropy.  
  (7) Optimizer: Adam.  
  
  (8) Learning Rate Scheduler: ReduceLROnPlateau or cosine annealing.  
  
  (9) Batch Size and Epochs: Tune for GPU memory; train for 20--30 epochs with early stopping.  
  
  (10) Hardware: Use GPU and multi-core CPU data loading. 
  
  (11) Output: Generate submission.csv strictly matching sample submission format.

  This iteration aims to exploit EfficientNetB4's superior feature extraction to significantly improve classification accuracy and reduce Multi-Class Log Loss.
  } \\

\end{longtable}
}

\newpage
{
\small
\setlength{\LTleft}{0pt}
\setlength{\LTright}{0pt}
\renewcommand{\arraystretch}{1.4}

\begin{longtable}{>{\RaggedRight}p{3.5cm} l >{\RaggedRight}p{7.5cm}}

  \caption{Instruction Dynamics of Qwen3-Coder-30B on MLE task \texttt{conways-reverse-game-of-life-2020}} \label{tab:instruction-dynamics-qwen3-coder-30b} \\
  \toprule
  \endfirsthead

  \toprule
  \endhead

  \midrule
  \multicolumn{3}{r@{}}{\textit{Continued on next page}} \\
  \endfoot

  \bottomrule
  \endlastfoot

  \multicolumn{3}{c}{\textit{Qwen3\_Coder\_30B\_Pretrain (score = 0.1543)}} \\ \midrule
  \multicolumn{3}{>{\RaggedRight}p{\linewidth}}{
  Implement a CNN-based model for Conway's Reverse Game of Life.
  
  The model should take the final state (25x25 grid) and delta steps as input and predict the initial state.
  
  Use PyTorch with CUDA support if available. Include data loading, model definition, training loop, and submission generation.
  
  The model should be designed to handle the toroidal boundary conditions properly.
  } \\
  \midrule

  \multicolumn{3}{c}{\textit{Qwen3\_Coder\_30B\_SFT (score = 0.1809)}} \\ \midrule
  \multicolumn{3}{>{\RaggedRight}p{\linewidth}}{
  Implement a Deep Learning solution using a Convolutional Neural Network (CNN) to reverse-engineer the initial state of Conway's Game of Life from a given final state. The goal is to predict the start state from the stop state and delta values.

  \textbf{Implementation Details:}
  
  (1) Data Loading and Preprocessing: Load train.csv and test.csv using pandas. Convert the start\_x and stop\_x columns into 25x25 NumPy arrays. Convert delta to a tensor. Normalize the arrays to binary values.  
  
  (2) Model Architecture: Design a CNN. Input is the concatenation of stop state and delta expanded to a 25x25 feature map. Use convolutional layers, batch normalization, and ReLU. Output is a 25x25 predicted start state. Clamp and round outputs to 0/1.  
  
  (3) Training: Use PyTorch Dataset and DataLoader, MSE loss, Adam optimizer, 10--20 epochs, optional early stopping or LR scheduling, CUDA support.
  
  (4) Inference: Predict on test.csv and generate submission.csv.  
  
  (5) Output Format: submission.csv contains id and 625 predicted start\_x values.  

  \textbf{Constraints and Best Practices:}
  
  Handle errors gracefully, optimize GPU usage, ensure correct tensor shapes, incorporate delta properly, and consider periodic boundary conditions.
  } \\
  \midrule

  \multicolumn{3}{c}{\textit{Qwen3\_Coder\_30B\_RL (score = 0.3777)}} \\ \midrule
  \multicolumn{3}{>{\RaggedRight}p{\linewidth}}{
  Implement a deep learning solution using a U-Net architecture to reverse Conway's Game of Life. This node focuses on debugging and establishing a robust baseline.

  \textbf{Implementation Details:}
  
  (1) Data Loading: Load train.csv and test.csv. Convert start and stop columns to (B,1,25,25). Expand delta to (B,1,1,1).  
  
  (2) Model Architecture: U-Net encoder-decoder with skip connections. Input is stop state concatenated with delta feature map. Final sigmoid layer outputs probabilities.  
  
  (3) Loss: MSELoss.  
  
  (4) Training: Adam optimizer (lr=0.001), batch size 32, 5 epochs, CUDA support. 
  
  (5) Inference: Threshold sigmoid outputs at 0.5 to obtain binary predictions. 
  
  (6) Submission: Format predictions into submission.csv.  
  
  (7) Code Structure: Self-contained script with data\_path and output\_path arguments.

  \textbf{Notes:}
  
  Delta ranges from 1 to 5. Dataset contains 50k examples. Periodic boundary conditions should be learned implicitly.
  } \\

\end{longtable}
}


\end{document}